\documentclass[letterpaper]{article}
\usepackage{aaai19}  
\usepackage{times}  
\usepackage{helvet}  
\usepackage{courier}  
\usepackage{url}  
\usepackage{graphicx}  

\usepackage{amsmath,amssymb,amsfonts}
\usepackage{algorithm}
\usepackage[noend]{algpseudocode}
\usepackage{textcomp}
\usepackage{xcolor}
\usepackage{amsthm}
\usepackage{mathtools}
\usepackage{upgreek}
\usepackage{booktabs}
\usepackage{subcaption}
\usepackage{xpatch}
\usepackage{colortbl}
\usepackage{bm}

\usepackage{tikz}
\usetikzlibrary{decorations.pathreplacing,calc}

\newcommand{\tikzmark}[2][-3pt]{\tikz[remember picture, overlay, baseline=-0.5ex]\node[#1](#2){};}

\tikzset{brace/.style={decorate, decoration={brace}},
	brace mirrored/.style={decorate, decoration={brace,mirror}},
}

\newcounter{brace}
\DeclareMathOperator*{\argmax}{arg\,max\,\,} 
\DeclareMathOperator*{\maxl}{max}

\newcommand{\est}{\text{est}}
\newcommand{\te}{\text{te}}
\newcommand{\tr}{\text{tr}}

\newcommand{\val}{\text{val}}

\newcommand{\mc}{\mathcal{X}}
\newcommand{\mct}{\mathcal{T}}

\newcommand{\commentout}[1]{%
}

\newcommand{\fs}{\fontsize{9.0pt}{10.0pt}}

\newcommand{\squishlist}{
	\begin{list}{$\bullet$}
		{ \setlength{\itemsep}{-.1ex}      \setlength{\parsep}{0ex}
			\setlength{\topsep}{0ex}       \setlength{\partopsep}{0ex}
			\setlength{\leftmargin}{.8em} \setlength{\labelwidth}{1em}
			\setlength{\labelsep}{0.5em} } }
	\newcommand{\squishend}{\end{list}}

\colorlet{tableheadcolor}{gray!25} 
\newcommand{\headcol}{\rowcolor{tableheadcolor}} %

\definecolor{Gray}{gray}{0.85}

\title{Learning Triggers for Heterogeneous Treatment Effects}
\author{Christopher Tran  \and Elena Zheleva\\
	Department of Computer Science, 
	University of Illinois at Chicago
	\\
	Chicago, IL, USA
	\\
	\{ctran29, ezheleva\}@uic.edu
}
\begin{document}
	
	\frenchspacing
	
	%
	
	\maketitle
	
	\begin{abstract}
		The causal effect of a treatment can vary from person to person based on their individual characteristics and predispositions. Mining for patterns of individual-level effect differences, a problem known as \emph{heterogeneous treatment effect estimation}, has many important applications, from precision medicine to recommender systems. In this paper we define and study a variant of this problem in which an individual-level threshold in treatment needs to be reached, in order to trigger an effect. 
		One of the main contributions of our work is that we do not only estimate heterogeneous treatment effects with fixed treatments but can also prescribe individualized treatments.
		We propose a tree-based learning method to find the heterogeneity in the treatment effects. Our experimental results on multiple datasets show that our approach can learn the triggers better than existing approaches. 
		
	\end{abstract}
	
	\section{Introduction}
	\label{sec:intro}
	
	Developing optimal precision treatments for diverse populations of interest can lead to more effective public policies~\cite{grimmer2017estimating}, medical decisions~\cite{laber2015tree,lakkaraju2017learning}, recommender systems~\cite{li2010contextual}, and more~\cite{ascarza2018retention}. 
	Treatment effects, also known as causal effects, assess the outcome response difference between applying a treatment to a unit and not applying one.
	\emph{Heterogeneous treatment effect (HTE) estimation} refers to finding subsets in a population of interest for which the causal effects are different from the effects of the population as a whole~\cite{athey2016recursive}. For example, if the treatment is a drug with potential adverse reactions, doctors may want to prescribe it only to those people who would benefit from it the most. Additionally, HTE analysis allows the discovery of subpopulations that can react adversely to a treatment and should avoid the treatment altogether. HTE can be studied as part of the post-analysis of running a controlled experiment or through observational data.
	
	Supervised machine learning techniques have been adapted to the problem of HTE estimation \cite{imai2013estimating,tian2014simple,xu2015regularized,grimmer2017estimating,xie2018false} and the closely related problem of finding individualized treatment regimes which aims to find the best individual treatment assignment \cite{almardini2015reduction,laber2015tree,lakkaraju2017learning,kallus2017recursive,kallus2018policy}. 
	Most of them rely on recursive partitioning using interpretable tree-based methods, such as decision lists \cite{lakkaraju2017learning}, decision trees \cite{athey2016recursive,breiman2017classification,laber2015tree,su2009subgroup,zeileis2008model} and random forests \cite{foster2011subgroup,wager2017estimation,athey2016generalized}. 
	Some of the splitting criteria include highest parameter instability \cite{zeileis2008model}, t-statistic for the test of no difference between splits \cite{su2009subgroup} and penalizing splits with high variance \cite{athey2016recursive}.
	In \cite{kallus2017recursive}, an impurity measure is developed to measure risk of treatments in a partition. There are also other methods that rely on clustering \cite{almardini2015reduction} or propensity scores \cite{xie2012estimating}.
	
	In many realistic scenarios, the treatment is an ordinal (or monotonously increasing continuous) variable, rather than a binary one, and the effect depends on the amount of treatment. For example, a clinician might be interested to understand the minimum number of days (the trigger) that patients with certain characteristics need to take a medication, in order to be cured (the effect). Or, a company might be interested in offering a personalized discount where the threshold is the minimum discount needed to trigger a customer with given characteristics to buy a product. Then, the goal becomes finding the threshold that maximizes the expected outcome for each discovered subpopulation where the subpopulation is characterized by its set of equal or similar attributes. To the best of our knowledge, none of the existing work addresses this problem which is the focus of our work.  
	
	We formalize this problem under the name \emph{trigger-based HTE estimation} and develop a learning procedure that enables the discovery of individual-level thresholds for triggering an effect.
	In essence, we turn an ordinal treatment problem into a binary treatment one where the treatment threshold is learned. For each subpopulation, the treatment effect refers to the average difference between the outcomes for the individuals in the subpopulation whose treatment is below the threshold and those who are above the threshold. A key assumption here is that the subpopulation that has received treatment above the threshold represents the same underlying distribution as the subpopulation below the threshold. Trigger-based HTE estimation is closely related to estimating dose-response curve used in medical fields for estimating effects of diet \cite{wang2014fruit} and survival analysis \cite{spratt2013long}. 
	In contrast, our method discovers subpopulations with heterogeneous triggers and effects.
	
	\begin{figure}[t]
		\centering
		\includegraphics[width=.30\textwidth]{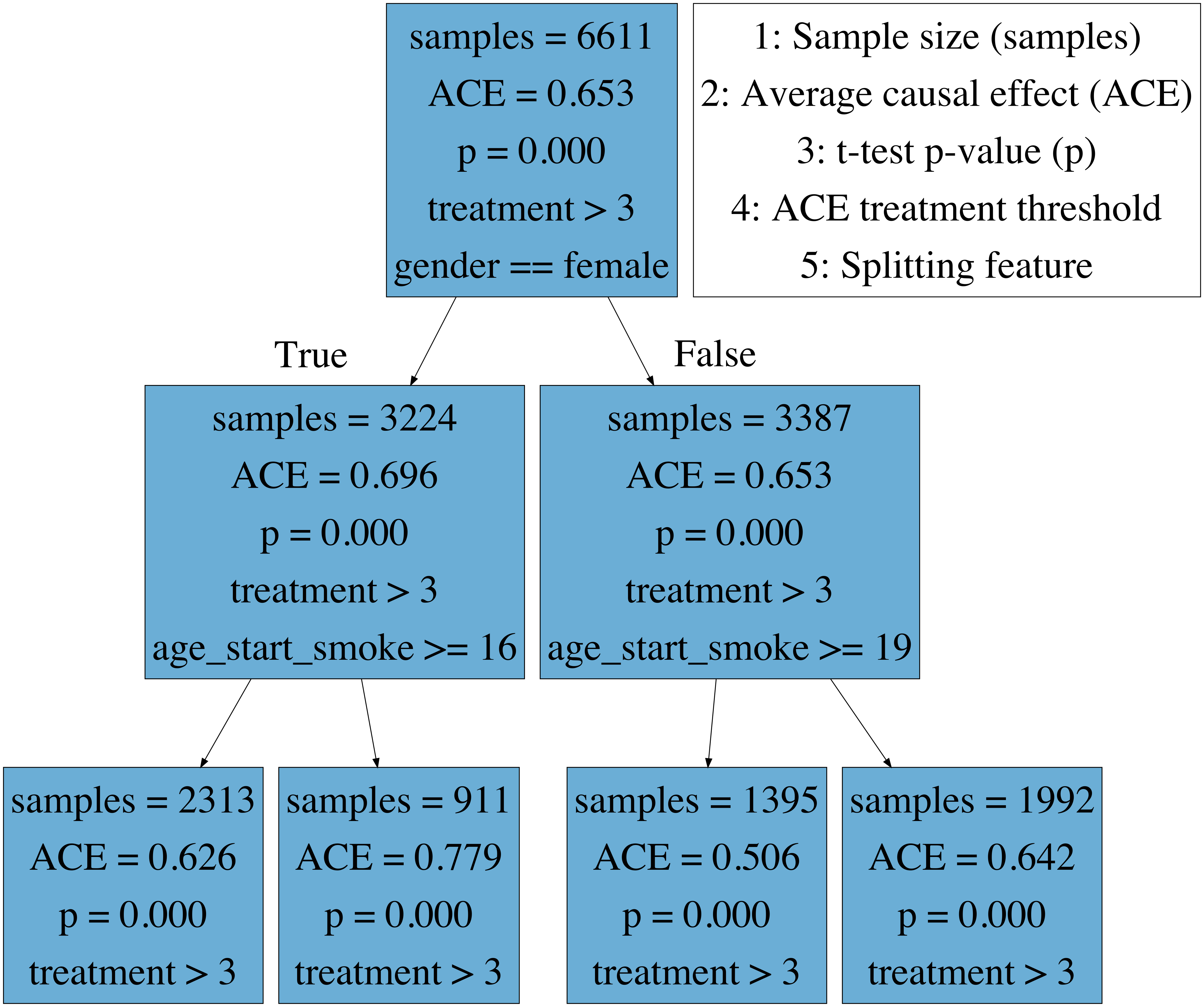}
		\caption{A small tree illustrating triggers for HTE. 
			The right most path reads: ``If a person is male, started smoking younger than 19", the average effect of smoking on medical expenditures is 0.642, for more than 3 $\log(\text{packyears})$. 
		}
		\label{fig:example-tree}
	\end{figure}
	
	Our proposed algorithm builds upon previous work on causal trees~\cite{athey2016recursive} which recursively partition a population using a decision tree-like approach. 
	There are two main challenges to learning trigger-based HTE that have not been addressed by previous work on causal trees. 
	First, a learning algorithm needs to be able to find effects that are robust and can generalize to unseen data. 
	Second, trigger-based HTE requires simultaneous learning of population-level triggers and population HTE.
	
	We propose a greedy algorithm with a novel scoring function that separates the full population into partitions with 
	different effects. To address the first challenge, we treat the causal effects in different possible populations as hyperparameters that need to be estimated from a validation dataset. Our node-splitting criterion learns to generalize to unseen data by introducing a penalty on a model's ability to estimate the conditional average causal effect. To address the second challenge, we search for treatment value thresholds that trigger the highest effects when binned into binary treatment and control groups based on that threshold. 
	
	Our main contributions include:
	\squishlist
	\item Framing a new problem of practical interest: trigger-based HTE estimation;
	\item Developing the first method for discovering groups with trigger-based HTE; 
	\item Utilizing trigger-based HTE estimation with causal trees for prescribing individualized treatments;
	\item Enabling causal effect generalization in causal tree learning through an appropriate node-splitting penalty; 
	\item Discovering HTE with significantly lower error rates compared to existing baselines. 
	\squishend
	
	To illustrate the trigger-based HTE groups that our method can discover, we show a simple example in Fig. \ref{fig:example-tree}. The data is from the 1987 National Medical Expenditures Survey (NMES) \cite{johnson2003disease}. 
	The treatment is amount of smoking (in $\log(packyears)$) and the effect is medical expenditures (in $\log(cost)$). 
	The first splitting feature is gender, where the left path is true (female) and the right path is false (male). 
	One example of a path in the tree is the right-most path: if a person is male and started smoking younger than 19 years old, then smoking more than $\log(packyears)$ leads to the highest difference in treatment effects compared to people with the same characteristics who smoked less, a $64.2\%$ increase in medical expenditures.

	\commentout{
		Recently, supervised machine learning techniques have been adapted to the problem of individual treatment effects~\cite{shalit2016estimating} and HTE~\cite{foster2011subgroup,tian2014simple,athey2016recursive}. A key principle in controlling complexity and generalization ability of machine learning models is the concept of using validation data for optimizing the model hyperparameters. The challenge to applying machine learning methods directly, is that the ``ground truth" for a causal effect is not observed. In other words, we observe a unit with treatment or without treatment, but not both at the same time. This is called the ``fundamental problem of causal inference"~\cite{holland1985statistics}. Therefore, it is not obvious how to utilize validation when building causal inference models.}
	
	\section{Preliminaries}
	\label{sec:prelim}
	
	First, we will present some background and definitions for causal inference, including assumptions needed for inferring treatment effects in observational data. Then, we present the heterogeneous treatment effect estimation problem. Finally, we briefly describe the causal tree approach proposed by~\cite{athey2016recursive}.
	
	\subsection{Causal Inference}
	\label{sec:causal-inference}
	
	Suppose dataset $S$ has $N$ units which are independently and identically distributed (i.i.d.). For a unit $s_i,  (i = 1, \dots, N)$, there are a pair of potential outcomes, $Y_i(0), Y_i(1)$. 
	The \emph{individual causal effect (ICE)} is defined as $\tau_i = Y_i(1) - Y_i(0)$, the difference in potential outcomes following the Rubin Causal Model~\cite{rubin1974estimating}
	Define the indicator for treatment group assignment as $T_i \in \{0, 1\}$, where $T_i = 0$ and $T_i=1$ indicates that unit $s_i$ is in the control or treatment group, respectively. Traditionally, we cannot observe the outcome when treated and outcome when not treated simultaneously.
	The actual observed outcome is defined as:
	\begin{equation} \label{eq:actual_outcome}
	\fs
	Y_i^{\text{obs}} = Y_i(T_i) = 
	\begin{cases}
	\fs
	Y_i(0), & \text{if $T_i = 0$,} \\
	Y_i(1), & \text{if $T_i = 1$.}
	\end{cases}
	\end{equation}
	We assume the potential outcome of one unit should not be affected by treatment assignment of other units (the stable unit treatment value assumption (SUTVA) \cite{rubin1978bayesian}). We maintain the assumption of \emph{unconfoundedness} \cite{rosenbaum1983central}, given as: $T_i \perp \left(Y_i(0), Y_i(1)\right) \mid X_i.
	$
	
	\subsection{Heterogeneous treatment effect estimation}
	\label{sec:hte}
	
	The main focus of heterogeneous causal inference is to estimate the \emph{conditional average treatment effect (CATE)} based on a set of features \cite{athey2015machine}. The CATE is defined as: $\tau(x) \equiv \mathbb{E} \left[ Y_i(1) - Y_i(0) | X_i= \bm{x} \right].$
	The goal is to obtain accurate estimates $\hat{\tau}(x)$ for the CATE $\tau(x)$, which are based on a partitioning of the feature space. 
	
	Consider the feature space $\mc \subset \mathbb{R}^d$ (we would like to note that we make this assumption for simplicity of exposition but our algorithms can easily be adapted for discrete features as well).
	A dataset $S$ consists of triples $S = \{(\bm{X}_i, Y_i, T_i) \colon \bm{X}_i \in \mc \}$ where $\bm{X}_i$ is the feature vector, $Y_i$ is the outcome or response, and $T_i$ is the binary indicator of treatment group assignment for unit $s_i$.
	A partitioning of the feature space into $L$ partitions is defined as
	$\mc = \mc_1 \cup \dots \cup \mc_L$, where $\mc_\ell \cap \mc_{\ell'} = \emptyset$ for all $\ell \neq \ell'$. 
	The set of examples corresponding to partition $\mc_\ell$ is $S_\ell = \{ (\bm{X}_i, Y_i, T_i) \colon X_i \in \mc_\ell \}$.
	
	The conditional mean for treatment and control in partition $\mc_\ell$ is defined as:
	\begin{equation}
	\fs
	\label{eq:mean}
	\hat{\mu}_t(S_\ell) = \frac{1}{N_{\ell_1}} \sum_{T_i = t, i\in S_\ell} Y_i,
	\end{equation}
	where $t\in\{0,1\}$, $\hat{\mu}_1$ and $\hat{\mu}_0$ are the conditional means for treatment and control groups in the partition, and $N_{\ell_1}$ and $N_{\ell_0}$ are the number of units in treatment and control groups in the partition, respectively. The \emph{average causal effect (ACE)}  $\hat{\tau}$ for partition $\mc_\ell$ is defined as:
	\begin{equation}
	\fs
	\label{eq:ace}
	\hat{\tau}(S_\ell) = \hat{\mu}_1(S_\ell) - \hat{\mu}_0(S_\ell),
	\end{equation}
	
	When performing estimation on a new set of data, say $S^\te$, the test examples are matched to the correct partition given the features. Given an example $s_j \in S^\te$ in partition $\ell$, the estimated ICE for unit $s_j$ is given by the ACE for $S_\ell$ as \eqref{eq:ace}.
	
	\subsection{Objective function}
	\label{sec:objective}
	
	The goal of HTE estimation is to partition the feature space, so that heterogeneity is found. Similar to~\cite{athey2016recursive}, we define a \emph{partition measure} that captures the magnitude in ACE for the partition: $F(S_\ell) = N_\ell \cdot \hat{\tau}^2(S_\ell)$,
	where $N_\ell = \lvert S_\ell \rvert$ is the number of samples in partition $\ell$.
	
	To find subgroups for HTE, we wish to
	partition the feature space, so that the sum of all partition measures is maximized.
	For a feature space $\mc$, 
	the \emph{objective} is to maximize the sum of partition measures $F$ across $L$ partitions:
	\fs
	\begin{equation}
	\begin{aligned}
	& \maxl_{S_1,\dots,S_L} & & \sum_{i=1}^L F(S_i), \\
	& \text{s.t.} & & \mc = \mc_1 \cup \dots \cup \mc_L, \\
	& & & \mc_\ell \cap \mc_{\ell'} = \emptyset, \; \ell \neq \ell'
	\end{aligned}	
	\label{eq:objective}
	\end{equation}
	\normalsize
	To reach this objective for the trigger-based HTE estimation, we need to optimize partition splits both based on features and possible feature threshold splits.
	
	\subsection{Recursive partitioning through causal trees} \label{sec:causal-tree}
	
	Trees are popular for HTE estimation~\cite{athey2016recursive,green2012modeling,laber2015tree}.
	Each partition is represented by a path in the tree, 
	similarly to decision trees. 
	Trees are built in a \emph{greedy} manner to maximize a $F$ from each node split. This results in locally optimizing the objective \eqref{eq:objective} as splitting criteria.
	Given a node $\ell$ that needs to be partitioned into two child nodes $\ell_1, \ell_2$, the algorithm finds the split that maximizes $F$ for the two children:
	\fs
	\begin{equation}
	\maxl_{S_{\ell_1},S_{\ell_2}} F(S_{\ell_1}) + F(S_{\ell_2})
	\label{eq:split}
	\end{equation}
	\normalsize
	such that $\mc_\ell = \mc_{\ell_1} \cup \mc_{\ell_2}$ and $\mc_{\ell_1} \cap \mc_{\ell_2} = \emptyset$.
	A simple method for building a tree is to directly use this splitting, called the \emph{adaptive} approach, which we denote as \textbf{CT-A}~\cite{athey2016recursive}. The entire training set is used to build the tree and estimate causal effects.
	
	A penalty can be introduced to the splitting criteria~\cite{athey2016recursive}.
	This approach, called the \emph{honest} approach, denoted as \textbf{CT-H}, is separated into \emph{two stages}: tree-building and estimation. The data is split into a \emph{training} and \emph{estimation} sample, denoted by $S^\tr$ and $S^\est$, respectively. $S^\tr$ is used for tree building and $S^\est$ is used for estimation.
	
	\subsection{Triggers for HTE}
	\label{sec:trigger-hte}
	
	When there is a treatment value threshold (e.g., minimum number of days to take a medication) that triggers an effect (e.g., to be cured), the partitions will depend on the optimal thresholds for each subpopulation. Let $t_i\in \mathbb{R}$ be the amount of treatment (e.g., number of days to take a medication) corresponding to each unit $s_i$. We define the \emph{trigger} $\theta_\ell$ of a population $S_\ell$ as the threshold that splits $S_\ell$ into two subpopulations $S_{\ell_1}$ and $S_{\ell_0}$ in a way that optimizes the treatment effect for $S_\ell$:
	\fs
	\begin{equation}
	\argmax_{\theta_\ell} \hat{\mu}_1(S_{\ell_1}) - \hat{\mu}_0(S_{\ell_0})
	\label{eq:trigger}
	\end{equation}
	\normalsize
	such that $S_\ell = S_{\ell_1} \cup S_{\ell_0}$, $S_{\ell_1} \cap S_{\ell_0} = \emptyset$, 
	where $T_i=1$ if $t_i\geq \theta_\ell$ and $T_i=0$ otherwise.
	While we define the treatment as a real variable, threshold-based triggers also apply to treatments that are ordinal discrete values. Like in the non-trigger case, a key assumption here is that $S_{\ell_1}$ and $S_{\ell_0}$ represent the same underlying distribution and each unit $s_i\in S_\ell$ is equally likely to be assigned to each subgroup ($T_i \perp \left(Y_i(0), Y_i(1)\right) \mid X_i$). 
	
	\section{Learning HTE}
	\label{sec:learning}
	
	Next, we present our general approach to learning HTE\footnote{Code available: https://github.com/chris-tran-16/CTL}, that applies to both discrete and trigger treatments. We explain in detail the specific requirements for the trigger case.
	
	\begin{algorithm}[t]
		\algrenewcommand\alglinenumber[1]{\small #1:}
		\small
		\caption{Learning trigger-based causal trees}
		\label{alg:learn-ct}
		\begin{algorithmic}[1]
			\Require Training set $S$, validation size $\rho$, indicator for measure $B$ ($B=1$ if binary, $B=0$ if trigger-based)
			\Ensure The root of the causal tree
			\State $S^\tr, S^\val$ = split($S$, split size=$\rho$)
			\State root.F $\gets$ 
			\Call{PartitionMeasure}{$S, B$} \{e.g. equation \eqref{eq:trigger-measure}\}
			\State root.S $\gets S^\tr$ \{The sample at root node\}
			\State \textbf{return} \Call{Train}{root}
			\Function{Train}{currentNode}
			\State $S_\ell \gets $ currentNode.S
			\State bestPartition $\gets -\infty$
			\For{each feature split $\mc_{\ell_1}, \mc_{\ell_2}$}
			\State leftPartition = \Call{PartitionMeasure}{$S_{\ell_1}, B$}
			\State rightPartition = \Call{PartitionMeasure}{$S_{\ell_2}, B$}
			\If{leftPartition + rightPartition $>$ bestPartition}
			\State bestPartition = leftPartition + rightPartition
			\EndIf
			\EndFor
			\If{bestPartition $>$ currentNode.F}
			\State left.F, left.S $\gets$ leftPartition, $S_{\ell_1}$
			\State right.F, right.S $\gets$ rightPartition, $S_{\ell_2}$
			\State currentNode.left $\gets$ \Call{Train}{left}
			\State currentNode.right $\gets$ \Call{Train}{right}
			\Else
			\State \textbf{return} currentNode
			\EndIf
			\EndFunction
			
			\Function{PartitionMeasure}{$S_\ell$, $B$}
			\If{$B == 1$ \{ indicates no trigger \}} 
			\State \textbf{return} $F_C(S_\ell)$
			\Else
			
			\State  $\theta_{\ell} \gets \emptyset$; \{denotes the trigger\}
			\State $v_\ell \gets -\infty$; \{$v$ is the best partition measure\}
			\State $\mct = \{t_i \colon t_i \in S_\ell\}$
			\For{each $t_j \in \mct$}   
			\State $S_{\ell_1} = \{(\bm{X}_i , Y_i, T_i) \colon T_i \geq t_j\}$
			\State $S_{\ell_0} = \{(\bm{X}_i , Y_i, T_i) \colon T_i < t_j\}$
			\State temp = $F(S_{\ell_1} \cup S_{\ell_0})$
			\If{temp $> v_\ell$}
			\State $v_\ell, \; \theta_{\ell}$ $\gets$ temp, $t_j$
			\EndIf
			\EndFor
			\EndIf
			\State \textbf{return} $v_\ell, \theta_\ell$
			\EndFunction
		\end{algorithmic}
	\end{algorithm}
	
	\subsection{Effect estimation}
	\label{sec:learning-hte}
	
	The formulation given previously may not be optimal on unseen test data.
	In contrast to \cite{athey2016recursive}, we propose a different splitting criterion by introducing the idea of a validation set to find splits that generalize well to unseen data. 
	For our approach, we clearly separate a training, validation, and testing sample for training and evaluation. 
	Additionally, we jointly optimize generalizability using training and validation in a \emph{one-stage} tree-building process.
	
	For a dataset $S$, we define the training, validation and testing samples as $S^\tr$, $S^\val$, and $S^\te$. We build the tree on the training portion, while penalizing generalization ability based on the validation set.
	Our method estimates effects on the training sample and penalizes estimations that do not match a validation set by introducing a penalty or \emph{cost}.
	
	Let $\Tilde{\tau}(S_\ell^\tr)$ be the true ACE in a node $\ell$.  The estimated ACE on the validation set is $\hat{\tau}(S_\ell^\val)$ as in \eqref{eq:ace}. Formally, define the \emph{cost term} $C$ as: $C(S_\ell^\val) = N^\val_\ell \cdot \left \lvert \hat{\tau}(S_\ell^\val) - \Tilde{\tau}(S_\ell ^\tr) \right \rvert$.
	This measures the error of estimated effect and validation ACE.
	Using the cost term $C$, define a new measure:
	\begin{equation}
	\fs
	\label{eq:learning-measure}
	F_C(S_\ell) = \frac{ (1-\lambda) \cdot F(S_\ell^\tr) - \lambda \cdot C(S_\ell^\val)}{\lvert N_\ell^\tr - N_\ell^\val \rvert + 1}.
	\end{equation}
	\normalsize
	where $\lambda \in [0, 1]$ is a parameter that controls the penalty. By adjusting $\lambda$, we can introduce higher penalty to splits that do not generalize well. 
	The denominator acts as normalization for training and validation sizes. 
	This gives fair comparison of measures across different splits. 
	This new measure maximizes partition measures, 
	but encourages generalization on unseen data through the penalty and the use of validation.
	
	The maximization problem for \emph{causal tree learning} is the same as in \eqref{eq:objective} using our new partition measure.
	We are finding heterogeneous partitions, while minimizing generalization error using a cost $C$. 
	This formulation is flexible, since any cost can be introduced. 
	We call this method \textbf{CT-L}.
	
	We propose two variants, \textbf{CT-HL} and \textbf{CT-HV}. This method uses the idea of honest estimation from \cite{athey2016recursive}, where a separate estimation set is used to penalize variance in the splits.
	We now introduce a separate estimation set as $S^\est$.
	Define an \emph{honest} term as $H(S_\ell)$:
	\begin{equation}
	\fs
	\label{eq:honest}
	H(S_\ell) = \left(  1 + \frac{N^\est}{N}  \right) \cdot \left(  \frac{V^{(1)} (S_{\ell})}{p} + \frac{V^{(0)} (S_{\ell})}{1-p}  \right),
	\end{equation}
	where $p=N^\tr/N$ is the share of treated units, and $V^{(1)} (S_{\ell})$ and $V^{(0)} (S_{\ell})$ are the variance of treated and control means.
	
	In the first variant, we combine the idea of honest estimation with our validation penalty. 
	We separate an estimation set from the training data to perform honest estimation to control variance. We also separate a validation set, so we have generalization cost and variance penalty. The partition measure is: $F_{HL}(S_\ell) = F_C(S_\ell) - H(S_\ell)$.
	We call this the \textbf{CT-HL} method.
	
	In the second variant, we do not separate an estimation set. We treat the validation as a hybrid estimation and validation set. Formally, the partition measure is: $F_{HV}(S_\ell) = F_C(S_\ell) - H_\val(S_\ell)$.
	$H_\val$ is the same as the honest penalty, except using the validation set instead of a separate estimation set.
	We denote this as the \textbf{CT-HV} method.
	
	\subsection{Learning triggers for HTE}
	\label{sec:learning-triggers}
	
	The goal of identifying triggers for HTE estimation is to find subpopulations in which both the trigger threshold and the individual characteristics play a role in the observed effect differences. 
	We define a partition measure that optimizes the treatment effect for each subpopulation through the choice of both the feature and the trigger threshold.  
	For each node $\ell$ and its corresponding sample $S_\ell$, we wish to find the trigger $\theta_\ell$ that maximizes the treatment effect in $S_\ell$. When finding splits, we jointly optimize for the choice of feature to split on and triggers of the node's children. 
	For example, consider estimating the effect that a product discount has on the decision to buy that product. In this case, the method would identify the minimum discount necessary to make a customer with certain characteristics buy that product.
	
	Let $\mct = \{t_i\}$ be the set of all possible treatment values in the dataset. Define $\theta_\ell \in \mct$ as the trigger value that splits the sample $S_\ell$ into two subsamples $S_{\ell_1}$ and $S_{\ell_2}$, found by maximizing the ACE as in \eqref{eq:trigger}. The partition measure for the trigger-based treatment effect is defined as $F^\mct$:
	\begin{equation}
	\fs
	\label{eq:trigger-measure}
	F^\mct(S_\ell) = \max_{\theta_\ell} \; F_C(S_\ell),
	\end{equation}
	where $S_\ell = S_{\ell_1} \cup S_{\ell_2}$ is the sample with trigger-based treatment. 
	We are maximizing the previously defined partition measure in \eqref{eq:learning-measure}, over the possible triggers. 
	For splitting, we use the trigger-based partition measure, replacing the binary partition measure in \eqref{eq:objective}. 
	Note that this formulation finds the best trigger for treatment at each node in the tree. 
	Therefore, the trigger for heterogeneity is different and can be observed at each level. 
	We note that the partition measure used in \eqref{eq:trigger-measure}, $F_C(S_\ell)$, can be any partition measure (e.g. $F_H(S_\ell)$). This makes our formulation easily applicable to any measure.
	
	Algorithm \ref{alg:learn-ct} briefly describes how to learn trigger-based causal trees. The algorithm requires the validation size, and an indicator for considering binary treatments or trigger-based treatments.
	To determine triggers, we consider all treatment values in the dataset as possible trigger values.
	
	One potential concern is our strong assumption that the population above and below the trigger represent the same underlying distribution. While decision trees take care of it to some extent (i.e., examples in a subpopulation stored in a leaf share the feature values of the tree path that leads to that leaf), in our experiments we ran further tests that compare the treatment and control population in each leaf.
	
	\section{Experiments}
	\label{sec:experiments}
	
	We compare our methods to the adaptive and honest causal trees developed in \cite{athey2016recursive}, as well as non-tree based methods that use propensity scores \cite{xie2012estimating}. 
	We study two datasets that lend themselves to the trigger-based treatment problem which is the focus of our work. We also use the ACIC Causal Inference Challenge dataset for binary treatments. 
	
	\begin{figure*}[ht]
		\centering
		\begin{subfigure}[b]{.40\textwidth}
			\includegraphics[width=\textwidth]{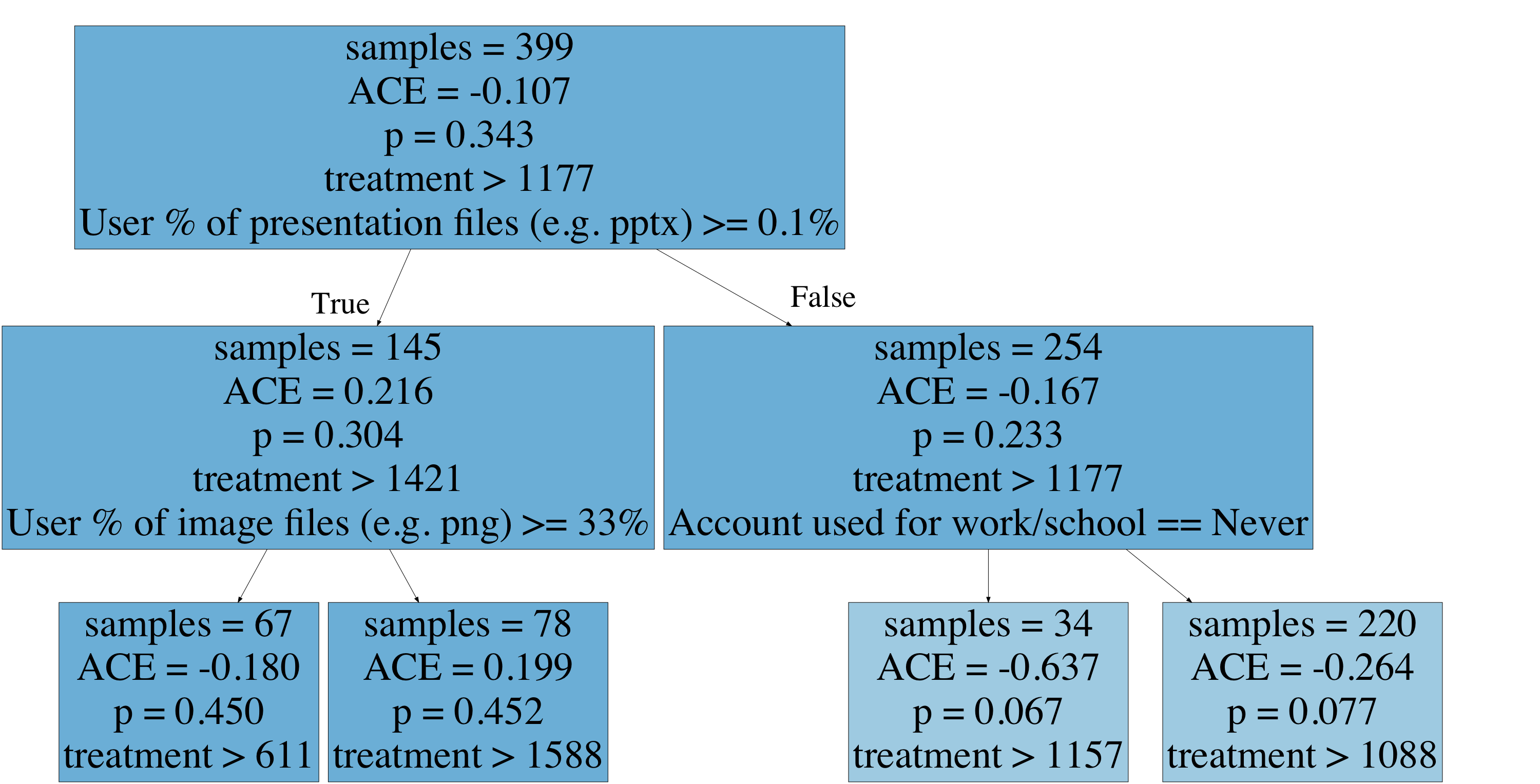}
			\caption{Trigger-based honest tree (CT-H)
				.}
			\label{fig:honest-cont}
		\end{subfigure}
		\begin{subfigure}[b]{.45\textwidth}
			\includegraphics[width=\textwidth]{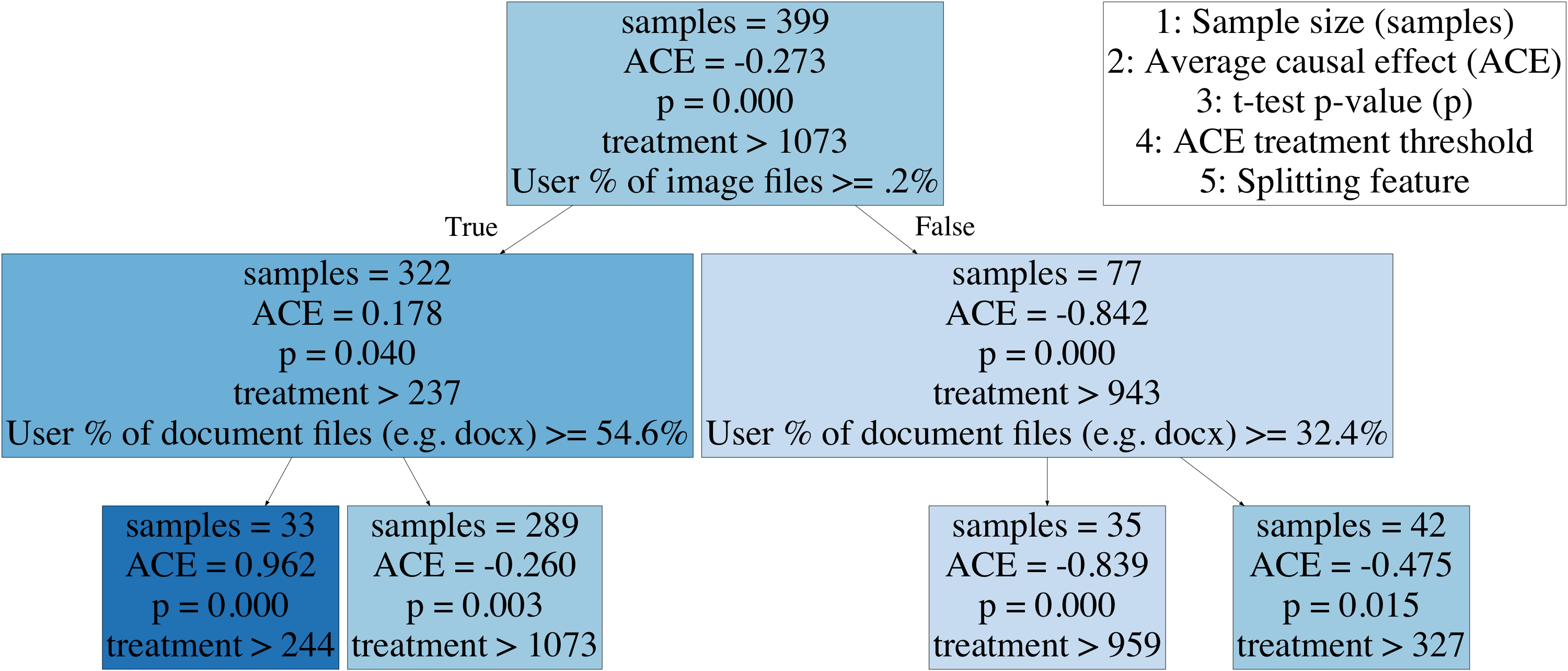}
			\caption{Our trigger-based tree (CT-L)
				.}
			\label{fig:learn-cont}
		\end{subfigure}
		\caption{Comparison between trigger-based CT-H and trigger-based CT-L on the cloud storage dataset.
			Darker and lighter shading indicate treatment effect is higher and lower, respectively.
			Our tree can find more statistically significant ACE's than the honest tree. 
			This suggests that the adaptation for the honest tree to triggers is not as effective. 
		}
		\label{fig:cont-compare}
	\end{figure*}
	
	\begin{table}[t]
		\centering
		\caption{Error for unpruned and pruned (e.g. CT-L vs CT-LP) trees on the trigger datasets. Bold results are statistically significant. Our methods (in grey) have lower error.
		}
		\label{tab:cont}
		\resizebox{.75\linewidth}{!}{
			\begin{tabular}{*{3} c}
				Method & NMES Error & Cloud Storage Error
				\\
				\toprule
				CT-A & 0.982 & 0.745 
				\tikzmark[xshift=.75em, yshift=-.25em]{u1}
				\\
				CT-H & 0.907 & 0.766
				\\
				\headcol CT-L & 0.597 & \bf0.607
				\\
				\headcol CT-HL & 0.417 & 0.613 
				\\
				\headcol CT-HV & \bf0.416 & 0.719 
				\tikzmark[xshift=.75em, yshift=-.25em]{u2}
				\\
				\midrule
				CT-AP & 0.982 & 0.699 
				\tikzmark[xshift=.75em, yshift=.33em]{p1}
				\\
				CT-HP & 0.894 & 0.667
				\\
				\headcol CT-LP & 0.566 & 0.573
				\\
				\headcol CT-HLP & 0.417 & 0.484
				\\
				\headcol CT-HVP & \bf0.416 & \bf0.396 
				\tikzmark[xshift=.75em, yshift=.33em]{p2}
				\\
			\end{tabular}
		}
	\end{table}
	
	\subsection{Experimental setup}
	\label{sec:exp-setup}
	
	We run experiments using the adaptive (CT-A) and honest (CT-H) trees~\cite{athey2016recursive}. We also compare to non-tree based methods that use propensity score such as stratification-multilevel (\textbf{SM}) method, matching-smoothing (\textbf{MS}) method, and smoothing-difference (\textbf{SD}) method \cite{xie2012estimating}. For propensity score estimation, we use logistic regression. 
	It is not obvious how to adapt the propensity based methods to the trigger-based problem, so we use them for the binary treatment evaluation only.
	For each dataset we do an 80-20 randomized split for the training and testing set for evaluation.
	For ours and honest trees, we split the training into two parts: a training $S^\tr$ and validation $S^\val$. For honest trees, $S^\val$ is the estimation set $S^\est$.
	
	In the experiments, we report results for pre-pruned and post-pruned trees. We grow a maximum sized tree and prune all trees based on statistical significance ($\alpha = 0.05$). For pruned trees, we report error only on leaf nodes that have statistically significant ACEs. These are denoted with a ``P" attached to the method (e.g. CT-HP is the pruned honest tree). We learn $\lambda$ and the validation split size for our methods on a separate validation set.
	
	\subsection{Datasets}
	\label{sec:datasets}
	
	We consider two types of datasets: datasets with continuous treatments used for the trigger-based treatment problem, and binary treatment datasets. We adapt CT-A and CT-H to the trigger-based treatment problem, by incorporating the trigger-based objective \eqref{eq:trigger} into their respective splitting criteria, and applying the same method for finding triggers.
	
	\textbf{Trigger treatments} For the first dataset, we use data from the 1987 National Medical Expenditure Survey (NMES) \cite{johnson2003disease}.
	We estimate the effect increased smoking has on medical expenditures. The treatment is the $\log({packyears})$ and the outcome is $\log({cost})$.
	
	The second dataset is a cloud storage usage survey, which asks users various questions about their cloud storage \cite{khan2018forgotten}. 
	We look at the effect age has on the file management decision. 
	Users were asked whether they would like to keep, encrypt, or delete the file. 
	We focus on the decision to keep or delete a file.
	
	\textbf{Binary treatments} The last dataset comes from the ACIC 2018 Causal Inference Challenge\footnote{https://www.cmu.edu/acic2018/data-challenge/index.html}. 
	The dataset provides 24 different simulations with sizes from 1000 to 50000.
	The dataset contains information about the treatment and the response of a sample. The pre-treatment and post-treatment response which provides ground-truth for evaluation is available. 
	We use this data as a proof our concept our methods on the binary case, but binary treatments are not our focus.
	
	\subsection{Evaluation}
	\label{sec:evaluation}
	
	For trigger-based evaluation (NMES and Cloud Storage), we report the symmetric mean absolute percentage error (SMAPE) for estimated and true ACE. For each leaf $\ell$, we get the predicted ACE using the estimator $\hat{\tau}(S_\ell^\te)$. The true ACE is the difference in population mean for the sample $S_\ell^\te$ as in \eqref{eq:ace}. 
	For $L$ leaves, the error for ACE is defined as:
	\begin{equation}
	\fs
	\label{eq:ace-error}
	\text{ACE\_Error}(S_\ell^\te) = \frac{1}{L} \sum_{\ell = 1}^{L} \frac{\left \lvert \hat{\tau}(S_\ell^\te) - \tilde{\tau}(S_\ell^\te) \right \rvert}{\lvert \hat{\tau}(S_\ell^\te) \rvert + \lvert \tilde{\tau}(S_\ell^\te) \rvert},
	\end{equation}
	
	For binary treatments (e.g. ACIC), we calculate SMAPE on the test set. Let the test dataset contain $N^\te$ examples. Define the true effect to be $\tau_i$ and the predicted effect to be $\hat{\tau}_i$. The SMAPE error on the test set is defined as:
	\begin{equation}
	\fs
	\label{eq:error}
	\text{Error}(\tau, \hat{\tau}) = \frac{1}{N^\te} \sum_{i=1}^{N^\te} \frac{\lvert \tau_i - \hat{\tau}_i \rvert } {\lvert \tau_i \rvert + \lvert \hat{\tau}_i \rvert}.
	\end{equation} 
	
	\begin{table}[t]
		\centering
		\caption{Variance across leaves to measure heterogeneity of effects. Our methods (in grey) have higher variance than the honest method. The adaptive method has the highest variance, but sacrifices lower error.}
		\label{tab:variance}
		\resizebox{.75\linewidth}{!}{
			\begin{tabular}{*{3}c}
				Method & NMES variance & Cloud variance
				\\
				\toprule
				CT-AP & 5.177 & 1.782
				\\
				CT-HP & 0.790 & 0.209
				\\
				\headcol
				CT-LP & 1.180 & 0.850
				\\
				\headcol
				CT-HLP & 0.974 & 0.243
				\\
				\headcol
				CT-HVP & 1.089 & 0.576
			\end{tabular}
		}
	\end{table}
	
	\begin{table}[t]
		\centering
		\caption{Trigger-based selection bias test using average Mahalanobis across leaves. Our methods (in grey) have generally lower distances.}
		\label{tab:mahalanobis}
		\resizebox{.87\linewidth}{!}{
			\begin{tabular}{*{3}c}
				Method & NMES Mahalanobis & Cloud Mahalanobis
				\\
				\toprule
				CT-AP & 1.378 & 1.979
				\\
				CT-HP & 0.496 & 0.833
				\\
				\headcol
				CT-LP & 1.652 & 0.593
				\\
				\headcol
				CT-HLP & 0.441 & 0.627
				\\
				\headcol
				CT-HVP & 0.261 & 0.404
			\end{tabular}
		}
	\end{table}
	
	\begin{figure*}[t]
		\centering
		\begin{subfigure}[t]{.25\textwidth}
			\includegraphics[width=\textwidth]{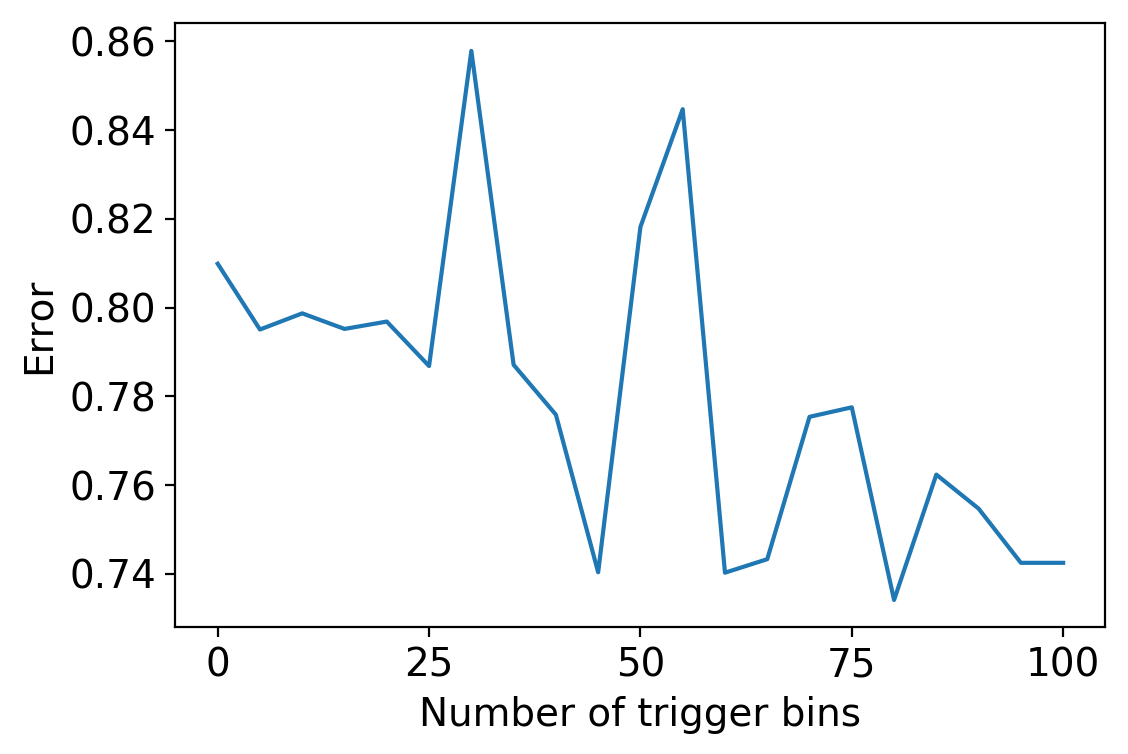}
			\caption{Number of triggers considered vs error on NMES data. 
			}
			\label{fig:discretize-plot}
		\end{subfigure}
		\hfill
		\begin{subfigure}[t]{.25\textwidth}
			\includegraphics[width=\textwidth]{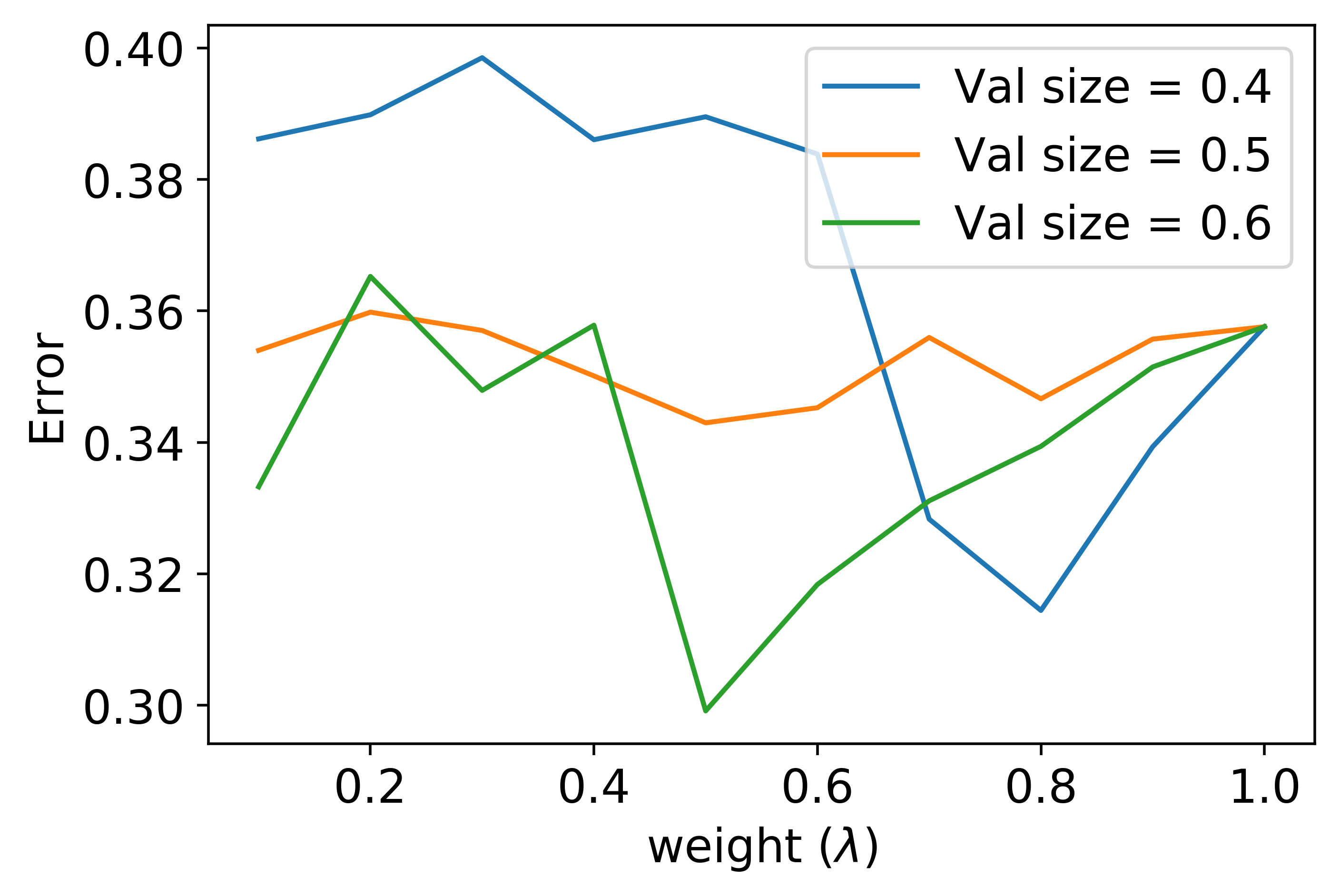}
			\caption{Error across with varying values of $\lambda$.}
			\label{fig:multi-weight}
		\end{subfigure}
		\hfill
		\begin{subfigure}[t]{.25\textwidth}
			\includegraphics[width=\textwidth]{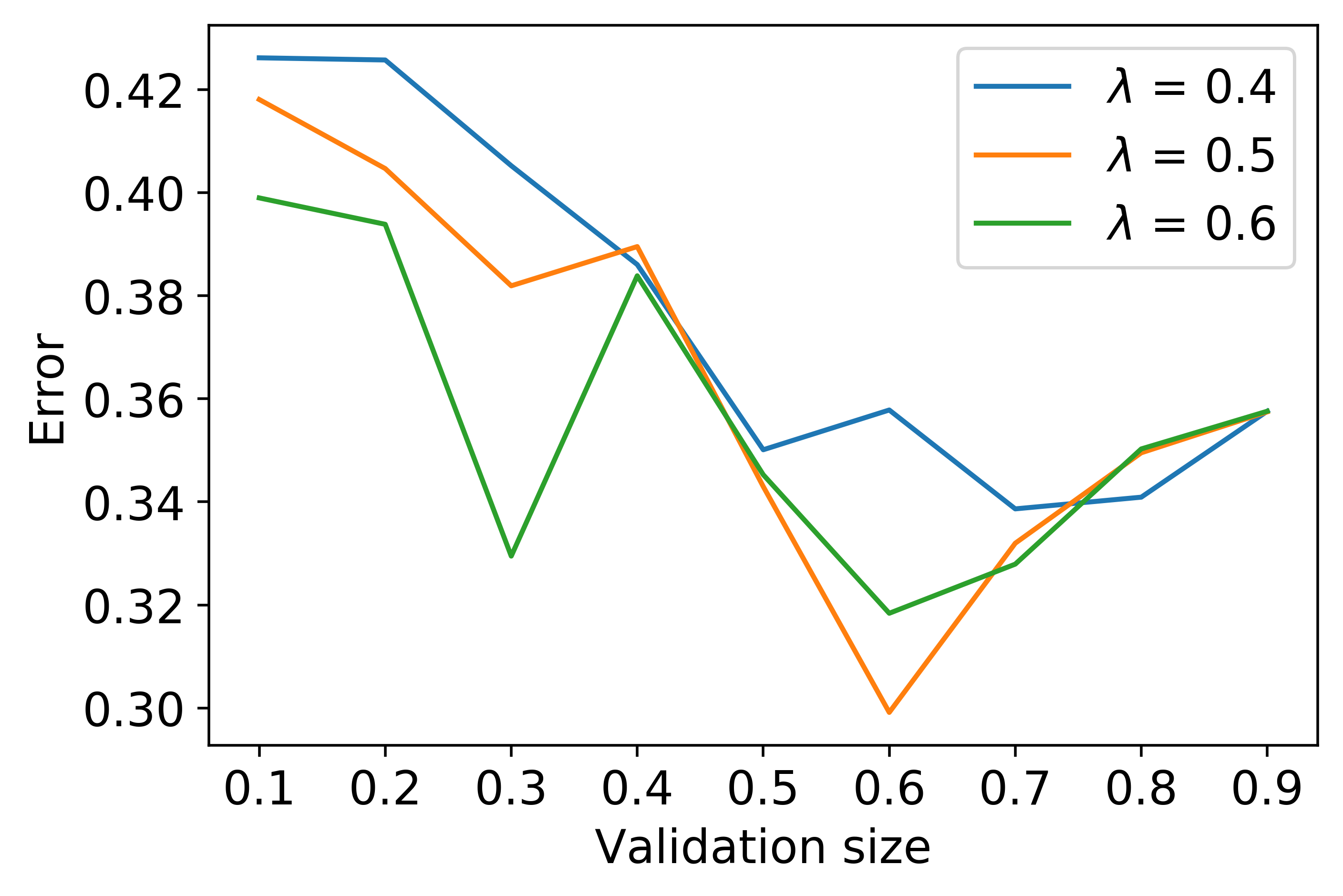}
			\caption{Error across with varying values of validation size.}
			\label{fig:multi-size}
		\end{subfigure}
		\hfill
		\caption{Plots showing effect of trigger discretization on error rate and hyperparameters using CT-HVP. 
			Fig. \ref{fig:discretize-plot} shows that considering more trigger values leads to lower error, but cannot compete with considering all triggers (0.741 compared to 0.416 in Table \ref{tab:cont}).
			Fig. \ref{fig:multi-weight} and \ref{fig:multi-size} show that varying values of $\lambda$ and the validation size effects the error, which suggests need for cross-validation. One way is to use the ACE\_error defined in \eqref{eq:ace-error}. 
		}
		\label{fig:three-figs}
	\end{figure*}
	
	\begin{table*}[t]
		\centering
		\small
		\caption{ACIC data error. The results are for non tree methods (e.g. MS), and the unpruned and pruned trees (e.g. CT-L vs CT-LP). 
			Bold results are statistically significant. The table shows that our methods (shaded grey), are significantly better in 18 out of 24 datasets. When our method does not perform better the other methods are not significantly better.}
		\label{tab:acic1}
		\resizebox{0.77\textwidth}{!}{
			\begin{tabular}{c *{4}c  *{4}c  *{4}c}
				& 
				\multicolumn{4}{c}{Data Size = 1000} & \multicolumn{4}{c}{Data size = 2500} & \multicolumn{4}{c}{Data size = 5000}
				\\
				Method 
				& 32be & 5316 & d09f & ea8e 
				& 6c04 & 7e4d & 95ba & c55e 
				& 4e47 & 9450 & a386 & f4c2
				\\
				\toprule
				MS
				& 0.922 & 0.897 & 0.685 & 0.895 
				& 0.828 & 0.823 & 0.872 & 0.663 
				& 0.914 & 0.895 & 0.117 & 0.200
				\\
				SD
				& 0.941 & 0.928 & 0.711 & 0.886 
				& 0.847 & 0.838 & 0.818 & 0.647 
				& 0.944 & 0.888 & 0.058 & 0.360
				\\
				SM
				& 0.826 & 0.998 & 0.400 & 0.882 
				& 0.864 & 0.953 & 0.710 & 0.191 
				& 0.864 & 0.995 & 0.997 & 0.223 
				\\
				\midrule
				CT-A 
				& 0.927 & 0.971 & 0.557 & 0.933 
				& 0.924 & 0.955 & 0.752 & 0.425 
				& 0.932 & 0.999 & 0.546 & 0.330
				\tikzmark[xshift=1em, yshift=0em]{au1}
				\\ 
				CT-H 
				& 0.970 & 0.998 & 0.473 & 0.831 
				& 0.905 & 0.891 & 0.985 & 0.244 
				& 0.860 & 0.998 & 0.855 & 0.220
				\\
				\headcol 
				CT-L 
				&
				0.958 & 0.946 & 0.319 & 0.876 & 
				\bf0.782 & 0.929 & \bf0.560 & 0.193 & 
				0.907 & 0.947 & 0.362 & 0.168
				\\
				\headcol 
				CT-HL 
				&
				0.827 & 0.980 & \bf0.189 & 0.926 & 
				0.807 & 0.894 & 0.626 & 0.185 & 
				0.802 & 0.999 & 0.297 & \bf0.016 
				\\
				\headcol 
				CT-HV 
				&
				0.935 & 0.986 & 0.296 & \bf0.790 & 
				0.793 & 0.887 & 0.826 & \bf0.127 & 
				0.909 & \bf0.794 & 0.679 & 0.108
				\tikzmark[xshift=1em, yshift=0em]{au2}
				\\
				\midrule
				CT-AP 
				& 0.893 & 0.971  & 0.484  & 0.978 
				& 0.920 & 0.954 & 0.756 & 0.247 
				& 0.930 & 0.999 & 0.433 & 0.290
				\tikzmark[xshift=1em, yshift=0em]{ap1}
				\\
				CT-HP 
				& 0.918 & 0.998 &  0.341 & 0.966
				& 0.905 & 0.961 & 0.985 & 0.105 
				& 0.924 & 0.998 & 0.855 & 0.149
				\\
				\headcol 
				CT-LP 
				&
				0.927 & 0.946 & 0.319 & 0.882 & 
				\bf0.782 & 0.924 & \bf0.524 & \bf0.054 & 
				0.918 & 0.947 & 0.193 & 0.135
				\\
				\headcol 
				CT-HLP 
				& 
				0.815 & 0.980 & \bf0.204 & 0.926 & 
				0.807 & 0.893 & 0.527 & 0.185 & 
				0.909 & 0.998 & \bf0.002 & \bf0.016 
				\\
				\headcol 
				CT-HVP 
				&
				0.926 & 0.986 & 0.296 & 0.889 & 
				0.801 & \bf0.762 & 0.826 & 0.076 & 
				0.944 & \bf0.883 & 0.003 & 0.043
				\tikzmark[xshift=1em, yshift=0em]{ap2}
				\\
				\bottomrule 
				\vspace{-6pt}
				\\
				& 
				\multicolumn{4}{c}{Data Size = 10000} & \multicolumn{4}{c}{Data size = 25000} & \multicolumn{4}{c}{Data size = 50000}
				\\
				Method 
				& 0099 & 0a2a & 5cc4 & c93b 
				& 2e47 & 4dce & 536d & 630b 
				& 3461 & 9d8c & a6c1 & f2e5
				\\
				\toprule
				MS
				& 0.649 & 0.914 & 0.657 & 0.398 
				& 0.830 & 0.845 & 0.754 & 0.867 
				& 0.764 & 0.918 & 0.256 & 0.910
				\\
				SD
				& 0.698 & 0.934 & 0.517 & 0.367 
				& 0.774 & 0.949 & 0.737 & 0.789 
				& 0.782 & 0.916 & 0.259 & 0.909 
				\\
				SM
				& 0.293 & 0.562 & 0.738 & 0.609 
				& 0.463 & 0.752 & 0.959 & 0.955 
				& 0.577 & 0.895 & 0.999 & 0.944 
				\\
				\midrule
				CT-A 
				& 0.197 & 0.872 & 0.142 & 0.605 
				& 0.350 & 0.872 & 0.457 & 0.961 
				& 0.998 & 0.947 & 0.996 & 0.939
				\tikzmark[xshift=1em, yshift=0em]{au3}
				\\
				CT-H 
				& 0.191 & 0.954 & 0.176 & 0.547 
				& 0.291 & 0.835 & 0.315 & 0.887 
				& 0.977 & 0.917 & 0.984 & 0.973
				\\
				\headcol 
				CT-L 
				&
				0.189 & 0.568 & 0.194 & 0.461 & 
				0.198 & \bf0.596 & 0.076 & 0.749 & 
				\bf0.658 & 0.894 & 0.920 & 0.712
				\\
				\headcol 
				CT-HL &
				0.251 & \bf0.156 & 0.177 & 0.657 & 
				0.178 & 0.632 & 0.153 & \bf0.589 & 
				0.909 & 0.998 & \bf0.002 & \bf0.016 
				\\
				\headcol 
				CT-HV &
				\bf0.092 & 0.753 & 0.185 & 0.349 & 
				\bf0.106 & 0.720 & \bf0.060 & 0.619 & 
				0.923 & 0.888 & 0.965 & 0.807
				\tikzmark[xshift=1em, yshift=0em]{au4}
				\\
				\midrule
				CT-AP 
				& 0.157 & 0.754 & 0.142 & 0.466 
				& 0.350 & 0.912 & 0.448 & 0.921 
				& 0.998 & 0.947 & 0.996 & 0.955
				\tikzmark[xshift=1em, yshift=0em]{ap3}
				\\
				CT-HP 
				& 0.084 & 0.706 & 0.176 & 0.366 
				& 0.291 & 0.885 & 0.108 & 0.878 
				& 0.977 & 0.915 & 0.983 & 0.848
				\\
				\headcol 
				CT-LP 
				&
				0.077 & 0.455 & 0.194 & 0.176 & 
				0.207 & 0.810 & 0.076 & 0.482 & 
				\bf0.658 & 0.894 & 0.882 & 0.714
				\\
				\headcol 
				CT-HLP
				&
				0.088 & \bf0.156 & 0.177 & 0.141 & 
				0.178 & 0.857 & 0.077 & 0.468 & 
				0.884 & 0.887 & \bf0.002 & \bf0.016
				\\
				\headcol 
				CT-HVP &
				\bf0.041 & 0.753 & 0.174 & \bf0.064 & 
				\bf0.106 & \bf0.720 & \bf0.060 & \bf0.451 & 
				0.923 & 0.888 & 0.965 & 0.807
				\tikzmark[xshift=1em, yshift=0em]{ap4}
				\\
				\bottomrule
			\end{tabular}
		}
	\end{table*}
	
	We compute the standard deviation of the SMAPE to get statistical significance on the difference between evaluation metrics for different methods.
	The result is statistically significant if the standard deviations do not intersect.
	
	\subsection{Evaluation on trigger treatments}
	\label{sec:results-trigger}
	
	\textbf{Error comparison}. Table \ref{tab:cont} shows results on the NMES dataset and cloud storage dataset. 
	The adaptive and honest trees were adjusted for building trees on the continuous treatment, 
	using their splitting criterion in tandem with the trigger-based measure. The results of our methods are shaded in grey.
	
	For the NMES dataset, we find that our methods give significantly lower error than the adaptive and honest approaches.
	Additionally, as seen in Fig. \ref{fig:example-tree}, our conclusion is consistent with \cite{imai2004causal} that prolonged smoking increases medical expenditure.
	Fig. \ref{fig:example-tree} show the gender and the age when people started smoking are discriminating factors for heterogeneity. 
	
	On the cloud storage dataset, we show that our method can get significantly lower error than the adapted CT-A and CT-H methods. Overall, pruning the trees for statistical significance improves the error. In the prepruned trees, our base method, CT-L, performs better. After pruning, CT-HV performs the best. From both datasets, we see that combining the honest penalty with our validation method and metric lowers the error for both trigger-based datasets.
	
	In Fig. \ref{fig:cont-compare}, we compare two small trees on the cloud storage dataset. 
	We wish to find the trigger age that changes file management decision.
	In this case, above and below the trigger is considered as old and new, respectively.
	We show the p-value based on a t-test for the average causal effect (ACE).
	Positive ACE means that older files are likely to be kept, while negative ACE means likely to be deleted.
	For example, in the root node for Fig. \ref{fig:learn-cont}, files newer than 1073 days are more likely to be kept.
	We also see our method can find partitions with ACE that are statistically significant in a low depth. 
	In contrast to the honest trees, statistically significant nodes are found on depth 2.
	This shows that the \emph{adaptation} of honest trees to trigger-based treatments is not as effective.
	
	\textbf{Heterogeneity comparison}. Here, we evaluate whether our proposed methods discover more heterogeneous populations by looking at the variance across the discovered leaf effects for each method, as compared to the estimated population effect. Table \ref{tab:variance} compares the variance across leaves for the tree-based methods. We observe that the adaptive method finds the highest variance or most heterogeneity. At the same time, the adaptive method has the highest error on both the NMES and Cloud storage datasets, suggesting that the discovered heterogeneity does not have good generalizability. In contrast, our methods consistently find more heterogeneous subgroups compared to the honest method, while having lower error on the datasets. A visual comparison can be made from Fig. \ref{fig:cont-compare}. We see that the causal effect estimations (ACE) vary much more across leaves in ours compared to CT-H.
	Visually, this is shown by the shading of nodes, where darker and lighter means more positive and negative ACE.
	The shading varies much more in CT-L.
	
	\textbf{Trigger-based selection bias}. Next, we test our assumption that the treatment and control populations in each leaf (with treatment value above and below the leaf trigger) have the same underlying distribution. Although trees take care of this to an extent (examples in the same leaf have the same features used on the path), we compare the distance of treatment and control groups. Table \ref{tab:mahalanobis} compares the average Mahalanobis distance for each tree-based method. For Mahalanobis distance, we compute the distance between each treatment feature vector to the mean of the control group, and each control feature vector the the mean of the treatment, and average over all distances in each leaf. From the table, our proposed CT-HV method has lower population differences at leaves, which means that it deals better with trigger-based selection bias.
	
	\textbf{Impact of number of triggers}. One concern of finding triggers is the increase in complexity of building causal trees. Instead of considering all possible treatment values in the dataset as triggers, we explore the impact of using lower number of possible triggers. Fig. \ref{fig:discretize-plot} shows the number of triggers against the error rate using CT-HV on the NMES dataset. 
	We see that increasing number of considered triggers generally lowers the error.
	However, we see that even after introducing 100 possible triggers, the error is still significantly higher than when we consider all possible triggers (0.741 compared to an error of 0.416 from Table \ref{tab:cont}). This shows that there is a significant tradeoff between number of triggers considered and error.

	\subsection{Evaluation on binary treatments}
	Table \ref{tab:acic1} shows the error for each ACIC dataset. For each dataset, we compare unpruned (e.g. CT-L) and pruned (e.g. CT-LP) tree results separately. We bold statistically significant results. Our methods are shaded in grey.
	
	Across all the datasets, we see that our proposed methods achieve lower error compared to the previous tree-based methods. For those datasets where our method is not significantly better, the other methods do not have significantly lower error. Our variants (CT-HL and CT-HV) generally have lower error than our base method. Careful validation should be done to choose the best method.
	
	We notice that some of the datasets have very large error error ($>0.9$). 
	A likely reason is that the range of the response variable varies wildly in some cases.
	For example, in dataset 9450, the range of response is about 429.
	When building the tree, large values affect the ACE at each node. 
	
	\subsection{Hyperparameter tuning}
	\label{sec:tuning}
	Since we introduce two hyperparameters in our methods, namely the weight $\lambda$ for controlling validation penalty, and the validation split size, we explore varying these values.
	
	Fig. \ref{fig:multi-weight} shows the error rate for varying values of $\lambda$ across different validation sizes. From this plot, we see that with sizes of the validation set, the lowest error changes with different values of $\lambda$. We notice that a larger $\lambda$ is needed for smaller validation, while a smaller $\lambda$ is needed for larger validation. Also, the error seems to be more stable when the validation is the same size as the training set.
	
	Fig. \ref{fig:multi-size} shows the error rate for varying validation sizes, with $\lambda$ fixed. We can infer that having a validation size close to the size of the training size is a good choice for lower error. From both figures, we conclude using a method for learning these parameters is important. Cross-validation can be used to tune these parameters (e.g. using \eqref{eq:ace-error}).
	
	\section{Conclusion}
	\label{sec:conclusion}
	
	In this paper, we define the problem of trigger-based heterogeneous treatment effect estimation. 
	To do this, we propose novel splitting criteria for building causal trees and incorporate validation data to improve generalizability. We introduce the use of a validation set for HTE estimation for learning optimal partitioning, by introducing a loss function that penalizes partitions whose effects do not generalize. 
	We showed our method can get better coverage on average causal effect on unseen test data when identifying triggers. 
	Our experimental results show our method performs better than previous causal tree methods for binary treatment estimation. 
	Important future directions include outlier detection for decreasing sample variance, developing methods for different distribution assumptions, such as non-i.i.d data, and incorporating matching methods into HTE estimation.
	
	\fontsize{9.0pt}{10.0pt} \selectfont
	\bibliography{AAAI-TranC.6544}
	\bibliographystyle{aaai}
	
\end{document}